\documentclass[11pt,twocolumn,twoside]{IEEEtran}
\usepackage{amsmath}
\usepackage{multirow}
\usepackage{tabularx}
\usepackage{graphicx}
\usepackage[caption=false]{subfig}
\usepackage[margin=0.75in,headheight=0.45in]{geometry}
\usepackage[pdftex]{epsfig}
\usepackage{amsfonts}
\usepackage{amssymb}
\usepackage{fancyhdr}

\newcommand{\eye}{\mathbb{I}}

\newcolumntype{L}[1]{>{\raggedright\arraybackslash}p{#1}}
\newcolumntype{C}[1]{>{\centering\arraybackslash}p{#1}}
\include{graphicsx}

\begin{document}
\title{\vspace{0.2in}\sc ClimAlign: Unsupervised statistical downscaling of climate variables via normalizing flows}
\author{Brian Groenke$^{1}$\thanks{$^1$University of Colorado, Boulder, $^2$Jupiter Intelligence Inc.}, Luke Madaus$^{2}$, Claire Monteleoni$^{1}$}

\maketitle
\thispagestyle{fancy}
\setcounter{footnote}{2}
\begin{abstract}
Downscaling is a common task in climate science and meteorology in which the goal is to use coarse scale, spatio-temporal data to infer values at finer scales. Statistical downscaling aims to approximate this task using statistical patterns gleaned from an existing dataset of downscaled values, often obtained from observations or physical models. In this work, we investigate the application of domain alignment to the task of statistical downscaling. We present ClimAlign, a novel method for \textit{unsupervised}, \textit{generative} downscaling using adaptations of recent work in \textit{normalizing flows} for \textit{variational inference}. We evaluate the viability of our method using several different metrics on two datasets consisting of daily temperature and precipitation values gridded at low ($1^\circ$ latitude/longitude) and high ($\frac{1}{4}^\circ$ and $\frac{1}{8}^\circ$) resolutions. We show that our method achieves comparable predictive performance to existing supervised statistical downscaling methods while simultaneously allowing for both conditional and unconditional sampling from the joint distribution over high and low resolution spatial fields. To the best of our knowledge, this is the first proposed method for unsupervised statistical downscaling, and one of very few proposed methods that allows for efficient sampling of synthetic data. We provide publicly accessible implementations of our method, as well as the baselines used for comparison, on GitHub\footnote{https://github.com/bgroenks96/generative-downscaling}.
\end{abstract}

\section{Motivation}

Global Climate Models (GCMs) provide valuable information about Earth’s past, present, and future climate. The predictions made by these models have a wide range of possible applications in the Earth sciences, as well as in engineering and risk analysis. However, climate models typically operate at very coarse spatial scales, often summarizing over one hundred square kilometers or more. Predictions at this scale are useful for global analyses but provide very limited local or regional information.

There has long been great interest in using this coarse scale spatial data to obtain predictions at finer scales that provide more localized information \cite{wilby1998statistical}. Such information could in turn be used for the prediction or risk assessment of extreme flooding, heat waves, or wildfires for the benefit of cities, businesses, and residents. Other applications could be in process control for major infrastructure like transportation, electricity, and water, or in geological surveys for forestry and land use.

Statistical downscaling aims to approximate the underlying climate processes using only statistical patterns learned from existing high resolution data, typically obtained from a regional climate model or gridded observation data. Bürger et al. \cite{burger2012downscaling} provide an overview of several traditional methods for statistical downscaling, including bias-correction spatial disaggregation (BCSD), quantile regression networks, and “expanded downscaling” (XSD).

Recent advances in machine learning and "big data" analysis have naturally led to interest in how such techniques can be applied to statistical downscaling. One common approach is to use machine learning algorithms or generalized linear models to learn \textit{pointwise} (i.e. one model per spatial location) estimators over the high resolution spatial field. Recent work by Vandal et al. \cite{vandal2017deepsd,vandal2018statistical} has shown some promise in applying super-resolution convolutional neural networks to the task of downscaling precipitation maps, introducing an adaptation called DeepSD. Other authors have proposed the use of super-resolution generative adversarial networks (e.g. ESR-GAN) for downscaling climate/weather variables \cite{singh2019downscalegan}.

One of the key challenges in statistical downscaling is the stochasticity inherent in the relationship between spatial scales. Even in cases when paired low and high resolution data are available, inconsistencies between the two scales are common. A climate model, for example, may predict a certain amount of rainfall over a large region, but both the magnitude and spatial distribution of the corresponding observed rainfall at that time and location may be significantly different. Furthermore, it is important to recognize that this problem is inherently \textit{under constrained}; i.e. there are many possible fine-scale realizations for the same low resolution constraint.

These challenges and limitations motivate our investigation of probabilistic methods for downscaling, i.e. given low resolution data $X$ and high resolution data $Y$, rather than learn a deterministic mapping $f: X \rightarrow Y$, we wish to learn a \textit{joint distribution} $p(X,Y)$ which enables us to then sample from the space of possible low and high resolution maps.

To date, most applications of machine learning methods to statistical downscaling have framed the problem as pointwise (or group) regression \cite{vandal2019intercomparison,sachindra2018statistical}, super-resolution \cite{vandal2017deepsd}, or direct maximum likelihood estimation of the high resolution data as a function of the low resolution data \cite{banoconfiguration}. Such methods, while often effective at producing approximations of the conditional mean, provide limited information about the underlying joint distribution over the two domains.

In this work \cite{groenkeclimalign}, we show how downscaling can be posed as a \textit{domain alignment} task between two random variables $X$ and $Y$ with some set of shared latent variables $\mathcal{Z}$. We propose the application of \textit{normalizing flows} for variational inference \cite{rezende2015variational} (see section \ref{flows}) to the problem of statistical downscaling. We then apply state-of-the-art methods from the field of deep, generative latent variable modeling to learn an \textit{unsupervised} downscaler, i.e. one that does not require the availability of paired training data.

\section{Preliminaries}

\subsection{Variational inference}
\label{vi}

Bayesian methods are valuable tools for statistical inference thanks to their ability to treat latent variables as drawn from a probability distribution, rather than being fixed estimates. Unfortunately, computational cost is commonly cited as a significant limitation of Bayesian inference. Variational inference provides an efficient method of approximating the posterior $p(Z | X)$ by constructing a lower bound on the marginal likelihood, $p(X)$, often referred to as the \textit{evidence lower bound} (ELBO). This is done using a known, tractable density $q(Z)$, called a \textit{variational} distribution. The goal is then to minimize the dissimilarity between $p(Z | X)$ and the variational distribution $q(Z)$.

\subsection{Normalizing flows}
\label{flows}

For highly complex posteriors, simple variational distributions like the spherical Gaussian will typically fail to provide an adequate approximation. Thus, there is a need for a more flexible class of variational densities.

Let $f: \mathbb{R}^d \rightarrow \mathbb{R}^d$ be a smooth, invertible mapping over a probability density $p(z)$ such that $z = f^{-1}(f(z))$. Then it follows that the resulting random variable $z' = f(z)$ has a density:
\begin{equation}
    p(z') = p(z)|\det{\mathbb{J}_{f^{-1}}(z')}| = p(z)|\det{\mathbb{J}_f(z)}|^{-1}
\label{eq:change_in_density}
\end{equation}
where $\mathbb{J}$ is the Jacobian, and the last equality follows from the inverse function theorem \cite{rezende2015variational}. The absolute value of the Jacobian determinant represents the change in density from applying $f$ to $z$. In general, the cost of computing the Jacobian is $\mathcal{O}(n^3)$. It is possible, however, to construct transformations where the Jacobian can be more efficiently computed.

Arbitrarily complex densities can then be constructed by chaining together a series of transformations $f_i$ and successively applying equation \ref{eq:change_in_density}:
\begin{equation}
    z_k = f_k \circ f_{k-1} \circ \cdots \circ f_1(z_0).
\end{equation}
Then the log density is given by:
\begin{equation}
    \log p_k(z_k) = \log p_0(z_0) - \sum_{i=1}^k{\log |\det{\mathbb{J}_{f_{i}}(z_{i-1})}}|.
    \label{eq:flow_density}
\end{equation}
The sequence of random variables $z_0, z_1, \dots, z_k$ is called a \textit{flow}. The sequence of corresponding distributions $p_0, p_1, \dots, p_k$ is called a \textit{normalizing flow} \cite{rezende2015variational}.

\section{Method}

\subsection{Downscaling as domain alignment}

Suppose $x\sim p_X$ and $y\sim p_Y$ are two related random variables over their respective domains $X$ and $Y$. The task of domain alignment is to construct a bijection $f: X \leftrightarrow Y$ which \textit{aligns} the two variables according to their shared structure, i.e. latent variables. One example of domain alignment in climate science is the task of \textit{climate analog mapping}, where the goal is to align climate states from different time periods.

In the context of downscaling, we let $X$ and $Y$ represent the domains of all possible low and high resolution spatial fields respectively, such that each sample $x \in X$ and $y \in Y$ represents a field of climate variables over a fixed geographic region. Furthermore, let $x \sim p_X$ and $y \sim p_Y$ represent the respective marginal distributions over random variables $x$ and $y$. Let $p_\text{XY}$ be the joint distribution over $x$ and $y$ such that $(x,y)_t \sim p_\text{XY}$ is a matching low/high resolution realization at time $t \in \mathcal{T}$. Assume the distributions $p_X$ and $p_Y$ are \textit{stationary} over the time range covered by $\mathcal{T}$. Then we propose to model $p_\text{XY}$ as:
\begin{equation}
    p_\text{XY}(x,y) = \int_{z\in \mathcal{Z}} p(x,y,z) dz
    \label{joint_dist}
\end{equation}
where $\mathcal{Z}$ represents a space of \textit{shared latent variables} between $x$ and $y$, and $p(x,y,z)$ is the full joint distribution over all three random variables $x$, $y$, and $z$. Unfortunately, the integral in $\ref{joint_dist}$ is generally intractable for high dimensional $\mathcal{Z}$. We consider instead the graphical model: $X \leftarrow \mathcal{Z} \rightarrow Y$. Under this model, the joint distribution in equation \ref{joint_dist} can be rewritten as:
\begin{equation}
    p(x,y,z) = p(x,y|z)p(z) = p(x|z)p(y|z)p(z)
    \label{eq:joint_model}
\end{equation}
where the last equality follows from the assumed conditional independence of $x$ and $y$ given $z$. Since the variables $z \in \mathcal{Z}$ are unobserved, we can choose a tractable prior density for $p(z)$, such as an isotropic Gaussian. We are then left with the problem of modeling the conditional distributions, $p(x|z)$ and $p(y|z)$. Observe that $p_X(x) = \int_{z\in\mathcal{Z}} p(x|z)p(z)dz$ (and equivalently for $p_Y(y)$). Thus, we can obtain a point mass estimate of $p(x|z)$ and $p(y|z)$ by constructing normalizing flows, $f_X: \mathcal{Z}\leftrightarrow X$ and $f_Y: \mathcal{Z}\leftrightarrow Y$. For an isotropic prior, $p(z)$, we can view $f_X$ and $f_Y$ as functions which map from a space with complex covariance structure (the observations) to one in which all spatial dimensions are independent from one another (the latent variables).

\subsection{Aligning climate variables}

We propose the application of \textit{AlignFlow} \cite{grover2019alignflow} to the task of aligning climate variables. AlignFlow attempts to learn the graphical model specified by equation \ref{eq:joint_model} using a pair of normalizing flows, one for each domain. AlignFlow uses a hybrid objective consisting of both maximum likelihood and adversarial losses\footnote{For background on adversarial learning, please see \cite{goodfellow2014generative}}:
\begin{multline}
    \mathcal{L}_{\text{AlignFlow}}(f_X, f_Y, c_X, c_Y;\lambda_X, \lambda_Y) = \\
    \mathcal{L}_{\text{GAN}}(c_X, f_X\circ f_Y^{-1}) + \mathcal{L}_{\text{GAN}}(c_Y, f_Y\circ f_X^{-1}) \\ - \lambda_x \mathcal{L}_{\text{MLE}}(f_X) - \lambda_y \mathcal{L}_{\text{MLE}}(f_Y)
\label{alignflow_obj}
\end{multline}
where $c_X$ and $c_Y$ are "critic" functions used to evaluate the adversarial loss, $\mathcal{L}_{\text{GAN}}$, and $\mathcal{L}_{\text{MLE}}$ is the log-likelihood of the respective marginal. Each critic can be viewed as a "judge" trained to differentiate between samples from the true marginals $p_X, p_Y$ and samples produced by the "generators", $f_X$ and $f_Y$. Hyperparameters $\lambda_x$ and $\lambda_y$ control the importance of the marginal likelihoods in the objective functions. The choice of $\lambda$ depends on the practitioner's desire for accurate density estimates versus perceptual quality. As $\lambda$ (i.e. either $\lambda_x$ or $\lambda_y$) approaches infinity, the objective becomes equivalent to standard maximum likelihood estimation. As $\lambda$ approaches zero, the objective becomes equivalent to adversarial training.

We pose statistical downscaling as a special case of domain alignment where the variables to be aligned are the low and high resolution spatial fields. Our approach for downscaling can be summarized as follows:
\begin{enumerate}
    \item Upsample (downscale) samples $x$ to match the dimensionality of samples from $Y$ using a simple, invertible projection (e.g. nearest neighbors interpolation).
    \item Choose a simple, multivariate prior $p_{\mathcal{Z}}$ over $z \in \mathcal{Z}$ such as a standard, isotropic Gaussian or logistic distribution. Note that the dimensionality of $\mathcal{Z}$ must match that of $X$ and $Y$.
    \item Construct \textbf{invertible} functions $f_X^{(\phi)}: \mathcal{Z} \leftrightarrow X$ and $f_Y^{(\psi)}: \mathcal{Z} \leftrightarrow Y$, where $\phi$ and $\psi$ are the trainable parameters for each transformation. Like Grover et al. \cite{grover2019alignflow}, we choose $f_X$ and $f_Y$ from the family of \textit{normalizing flows} (see section \ref{flows}).
    \item Construct critic functions $c_X^{(\theta)}: X \rightarrow \mathbb{R}$ and $c_Y^{(\pi)}: Y \rightarrow \mathbb{R}$ to evaluate the adversarial loss, where $\theta$ and $\pi$ are the trainable parameters for $c_X$ and $c_Y$ respectively.
    \item Iteratively update the parameters for $f_X^{(\phi)}, f_Y^{(\psi)}, c_X^{(\theta)}, c_Y^{(\pi)}$ using stochastic gradient descent on the AlignFlow objective given in equation \ref{alignflow_obj}. This is done by independently sampling random (unpaired) batches $x\sim X$ and $y \sim Y$ from the training data. For each batch of samples, we obtain $\hat{y} = f_Y \circ f_X^{-1}(x)$ and $\hat{x} = f_X \circ f_Y^{-1}(y)$\footnote{Parameter superscripts are suppressed for brevity}. The log likelihood and adversarial losses are then computed and backpropagated using the predictions $\hat{y}$ and $\hat{x}$.
\end{enumerate}

Log likelihoods are computed according to equation \ref{eq:flow_density}. The adversarial objective is chosen to be the gradient penalized Wasserstein distance as described by Gulrajani et al. \cite{gulrajani2017improved} Unlike Grover et al. \cite{grover2019alignflow}, we construct $f_X$ and $f_Y$ as Glow \cite{kingma2018glow} normalizing flows due to their well demonstrated ability to generate high quality images. For the critics, $c_X$ and $c_Y$, we use the PatchGAN discriminator architecture \cite{isola2017image}. More details on the architecture and approach can be found in \cite{groenkeclimalign}. We refer to our method hereafter as \textit{ClimAlign}.

\subsection{Sampling}

AlignFlow permits both \textit{conditional} and \textit{unconditional} sampling via the tractable prior density, $p_Z$. Unconditional sampling from the joint distribution $p_{XY}$ is performed by drawing samples $\tilde{z} \sim p_{\mathcal{Z}}$ and evaluating $\tilde{x} = f_X(\tilde{z})$ and $\tilde{y} = f_Y(\tilde{z})$. We can interpret such samples $(\tilde{x},\tilde{y})$ from ClimAlign as low/high resolution maps drawn from the distribution over all possible spatial realizations of the climate variable, marginalized over time.

Conditional sampling is performed by evaluating the cross-domain mapping, $\hat{y} = f_Y\circ f_X^{-1}(x)$ or $\hat{x} = f_X\circ f_Y^{-1}(y)$ given some existing datapoints, $x$ or $y$. However, for a fixed datapoint and fixed parameters $\phi,\psi$ of $f_X^{(\phi)}, f_Y^{(\psi)}$, the mapping is deterministic. As previously discussed, we are interested in a \textit{probabilistic} mapping from which many high resolution realizations could be sampled given a fixed, low resolution input. Let $p(y|x)$ be the desired conditional distribution induced by such a mapping between $X$ and $Y$. Then $p(y|x)$ can be represented as:
\begin{equation}
p(y|x) = \int_\epsilon p(y|x,\epsilon)p(\epsilon)d\epsilon
\end{equation}
where $\epsilon \sim q(0,\sigma\eye)$ and $q$ is from the same family of distributions as $p_{\mathcal{Z}}$ (e.g. an isotropic Gaussian). For a fixed input $x$ and sampling temperature $\sigma$, conditional samples can then be computed as:
\begin{equation}
    \tilde{y} = f_Y(f_X^{-1}(x) + \epsilon)
\end{equation}
The output samples $\tilde{y}$ are effectively the result of making small, random perturbations to the latent vector representation of the input, $z = f_X^{-1}(x)$. Since $f_X$ and $f_Y$ are both smooth, invertible mappings, we expect that vectors in the neighborhood of $z\in \mathcal{Z}$ should be mapped to similar (i.e. spatially correlated) outputs in $X$ and $Y$ space. The sampling temperature $\sigma$ controls how much $z$ is perturbed, and consequently, how much the outputs will vary from the expected value of $p(y|x)$, where $\epsilon = 0$.

\section{Evaluation}

\begin{table*}[t!]
    \centering
    \caption{Test statistics, ERA-I $\rightarrow$ WRF-4; Daily max temperature}
    \begin{tabular}{| c | c | c c c | c c |}
    \hline
     Region & Method & RMSE ($^{\circ}$C) & Bias ($^{\circ}$C) & Corr & TXx$^\dag$ ($^{\circ}$C) & TXn$^\dag$ ($^{\circ}$C) \\
     \hline
     \multirow{3}{*}{SE-US}
     & BCSD & 1.51 $\pm$ 0.15 & -0.02 $\pm$ 0.21 & \textbf{0.93} $\pm$ 0.05 & -0.23 $\pm$ 0.93 & 0.14 $\pm$ 1.42 \\  
     & BMD-CNN & \textbf{1.30} $\pm$ 0.12 & 0.03 $\pm$ 0.13 & 0.90 $\pm$ 0.05 & -0.56 $\pm$ 0.79 & 0.74 $\pm$ 1.28 \\
     & ClimAlign (ours) & 1.56 $\pm$ 0.13 & \textbf{-0.005} $\pm$ 0.22 & 0.87 $\pm$ 0.06 & \textbf{0.003} $\pm$ 1.0 & \textbf{-0.04} $\pm$ 1.53 \\
     \hline
     \multirow{3}{*}{P-NW}
     & BCSD & 1.54 $\pm$ 0.23 & \textbf{0.01} $\pm$ 0.10 & \textbf{0.95} $\pm$ 0.03 & -0.23 $\pm$ 0.93 & \textbf{0.13} $\pm$ 1.42 \\  
     & BMD-CNN & \textbf{1.25} $\pm$ 0.14 & -0.06 $\pm$ 0.05 & 0.93 $\pm$ 0.02 & -0.73 $\pm$ 1.07 & 0.46 $\pm$ 1.04 \\
     & ClimAlign (ours) & 1.58 $\pm$ 0.18 & 0.03 $\pm$ 0.15 & 0.89 $\pm$ 0.04 & \textbf{0.22} $\pm$ 1.31 & 0.26 $\pm$ 1.46 \\
     \hline
    \end{tabular}\\
    $^\dag$\footnotesize{Standard deviations in these columns are over all months in the test set, not folds}
    \label{tab:stat_maxt}
\end{table*}

\begin{table*}[t]
\centering
\caption{Test statistics, ERA-I $\rightarrow$ WRF-4; Daily precipitation}
\begin{tabular}{| c | c | c c c | c c c c c |}
\hline
 Region & Method & RMSE (mm) & Bias (mm) & Corr & CDD & CWD & Rx1d & Rx5d & SDII \\
 \hline
 \multirow{3}{*}{SE-US}
 & BCSD & 27.32 $\pm$ 5.0 & 0.95 $\pm$ 1.4 & 0.39 $\pm$ 0.07 & 0.56 & \textbf{0.56} & 0.33 & 0.46 & 0.32 \\  
 & BMD-CNN & \textbf{14.11} $\pm$ 2.18 & -0.23 $\pm$ 0.47 & \textbf{0.50} $\pm$ 0.10 & \textbf{0.60} & 0.53 & \textbf{0.41} & \textbf{0.68} & \textbf{0.41} \\
 & ClimAlign (ours) & 18.40 $\pm$ 2.64 & \textbf{0.08} $\pm$ 0.86 & 0.42 $\pm$ 0.07 & 0.50 & 0.52 & 0.38 & 0.54 & 0.32 \\
 \hline
 \multirow{3}{*}{P-NW}
 & BCSD & 8.90 $\pm$ 2.30 & 0.41 $\pm$ 0.26 & 0.61 $\pm$ 0.06 & 0.80 & \textbf{0.70} & 0.56 & 0.70 & 0.58 \\  
 & BMD-CNN & \textbf{5.77} $\pm$ 0.72 & \textbf{-0.18} $\pm$ 0.61 & \textbf{0.70} $\pm$ 0.03 & \textbf{0.82} & \textbf{0.70} & \textbf{0.73} & \textbf{0.84} & \textbf{0.73} \\
 & ClimAlign (ours) & 7.33 $\pm$ 0.69 & 0.54 $\pm$ 0.54 & 0.67 $\pm$ 0.03 & 0.73 & 0.63 & 0.70 & 0.79 & 0.66 \\
 \hline
\end{tabular}
\label{tab:stat_prcp}
\end{table*}

We compare ClimAlign to existing downscaling methods using a variety of metrics on the ERA-interim and Rasmussen/Liu et al. WRF datasets. Both datasets consists of 4748 daily maximum temperature and precipitation values (approximately 13 years) over the continental United States (CONUS). High resolution predictions are obtained from the Weather Research and Forecasting model (WRF), which was forced using ERA-interim (ERA-I) \cite{berrisford2009era} as the boundary conditions \cite{liu2017continental}. The native resolution for WRF is roughly $\frac{1}{32}^{\circ}$ (in latitude/longitude) while ERA-interim is close to $0.7$ degrees. For computational convenience, ERA-interim is interpolated to 1 degree, while WRF is upscaled via bilinear interpolation to $\frac{1}{4}^\circ$. For our downscaling task, we use ERA-I as low resolution inputs to the downscaling algorithms. We use the $\frac{1}{4}^\circ$ upscaled WRF dataset (hereafter referred to as WRF-4) as our high resolution target.

We use two existing downscaling methods as baselines: \textit{Bias correction spatial disaggregation} \cite{wood2002long,vandal2019intercomparison} (BCSD)
and a deep convolutional inference network proposed by Ba\~no-Medina et al.\footnote{Specifically, we use CNN-10 as described in \cite{banoconfiguration}} \cite{banoconfiguration} (BMD-CNN). BCSD is a simple quantile mapping/linear interpolation technique widely used in the climate science community due to its simplicity and effectiveness. BMD-CNN is a more recently proposed application of deep neural networks to the task of downscaling. Both methods are \textit{supervised} (i.e. they require paired low/high resolution images), but they serve as a reasonable basis for comparison to test the viability of ClimAlign.

Predictive downscaling performance is assessed using both pointwise error metrics and a subset of the Climdex indices \cite{alexander2011climdex} for temperature and precipitation extremes. For each method, the root mean squared error (RMSE), bias, and Pearson's R (correlation) are reported. Metrics are evaluated per grid location between paired low/high resolution test samples and averaged over both time and space, unless otherwise specified. For precipitation, \textit{sparse} RMSE and bias are used in order to adjust for the high frequency of zeros in precipitation data.. We propose a sparse error metric to be:
\begin{equation}
    \frac{1}{N - N_0}\sum_t f(y_{ij}^{(t)}, \hat{y}_{ij}^{(t)})
\end{equation}
where $N_0 = \sum_t (1-\delta_{\epsilon}(y_{ij}^{(t)}))\times (1-\delta_{\epsilon}(\hat{y}_{ij}^{(t)}))$ is the number of days where both the true and predicted precipitation values are less than the aforementioned precipitation threshold, $\epsilon$. Note that here $\delta_{\epsilon}(.)$ is an indicator function and $f$ is the respective error metric (RMSE or bias).

Climdex indices are computed for both the predictions and observations. For daily max temperature, we report the bias between the predicted and observed \textit{monthly maximum} (TXx) and the \textit{monthly minimum} (TXn) values. For daily precipitation, we report the correlation between the predicted and observed \textit{maximum monthly 1-day} and \textit{5-day} precipitation (Rx1d, Rx5d), the \textit{longest number of consecutive "wet" days} (CWD) and \textit{"dry" days} (CDD) per month, and the \textit{simple precipitation intensity index} (SDII), which measures the average amount of precipitation on all "wet" days in any given month. We choose correlation instead of bias for precipitation in order to provide a less noisy measurement of the estimators' skill.

Each model is tested using 5-fold time series cross validation on a holdout test set consisting of the last 730 days (2 years) of the ERA-I/WRF-4 datasets. Each subsequent fold uses the next 146 days of the test set for validation and a sliding window of the last 4018 days ($\approx 11$ years) for training. Pointwise error metrics, averaged over time and space, are reported for each fold along with the standard deviation between folds. Since the Climdex indices are monthly statistics, a test set of 146 days would provide only 5 data points per fold, making test statistics unreliable. For robustness, Climdex indices are instead computed over the full test set. Experiments were repeated for both the Southeast US (SE-US) and Pacific NW (P-NW) regions.

\subsection{Daily predictions}

It should be noted that we do not generally expect unsupervised methods to outperform supervised methods in pointwise error for predictive tasks such as downscaling. However, the loss of per-pixel predictive power is offset by the model's sampling and density estimation capabilities. Furthermore, curating "labels" (or pairs) for supervised learning is often expensive or impossible in some contexts. Thus, if we can obtain similar predictive performance without such requirements, then this could be considered to be a worthwhile trade-off.

Table \ref{tab:stat_maxt} shows the comparative performance of each downscaling method on the ERA-I/WRF-4 datasets (4x resolution). ClimAlign achieves competitive results to both supervised methods on all metrics despite the lack of supervision during training. Additionally, ClimAlign outperforms the other methods with respect to monthly bias for both Climdex indices in the Southeast US region and for maximum monthly temperature (TXx) in the Pacific northwest. It does, however, tend to have greater variance across months.

Table \ref{tab:stat_prcp} shows the results of the same experiment but repeated for daily precipitation, which generally poses a more difficult task for statistical downscaling. This is immediately apparent by the significantly weaker correlation scores for all three methods' predictions. Overall, ClimAlign outperforms BCSD on most daily prediction statistics for both regions. BMD-CNN, however, achieves superior performance on nearly all metrics.

ClimAlign outperforms BCSD on both monthly rainfall statistics (Rx1d and Rx5d) as well as on intensity (SDII). ClimAlign shows inferior performance, however, on both occurrence statistics (CDD/CWD), which seems to indicate that it has difficulty in predicting the frequency of precipitation events. BMD-CNN obtains the highest correlation on almost all Climdex metrics for both regions. All methods show better predictive error and monthly Climdex correlations on the Pacific northwest region compared to the Southeast US, thus demonstrating the significance of geographic location in the effectiveness of statistical downscaling. This discrepancy may be due to more variance in the occurrence and magnitude of rainfall in the Southeast US region.

\begin{figure}[h!]
    \centering
    \includegraphics[width=\linewidth]{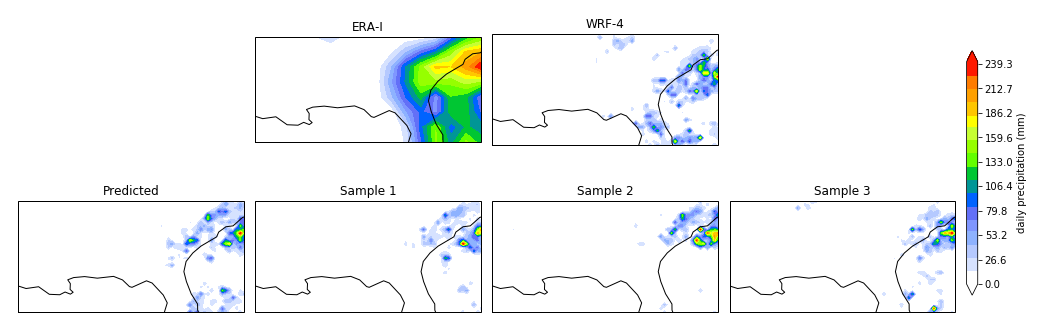}
    \caption{
    ClimAlign model outputs for ERA-I $\rightarrow$ WRF-4, Daily precipitation; Example input and target (top) Maximum likelihood prediction and conditional samples, $\sigma=0.7$ (bottom)
    }
    \label{fig:sample_prcp}
\end{figure}

Figure \ref{fig:sample_prcp} shows the true values for ERA-interim and WRF-4 (top row) along with the maximum likelihood prediction and random samples conditioned on the input (bottom row) for a randomly selected day in the test period\footnote{For all qualitative results, random selection was intentionally biased towards days with higher total precipitation.}. While each conditional sample is still similar to the maximum likelihood prediction, there appears to be a realistic amount of variance across the samples in where the precipitation events occur.

We next follow Grover et al. \cite{grover2019alignflow} in studying the structure of the latent space learned by ClimAlign. For this experiment, two "real" high resolution samples, $y_1$ and $y_2$, are chosen at random from the WRF-4 dataset and encoded using the trained model to obtain $z_1 = f_Y^{-1}(y_1)$ and $z_2 = f_Y^{-1}(y_2)$. Spherical linear interpolation (also known as \textit{Slerp}) is used to interpolate between $z_1$ and $z_2$. Slerp interpolates along a spherical path which has been found to better follow the manifold of learned examples in latent Gaussian models \cite{white2016sampling}. The interpolation function is defined as:
\begin{equation}
    \text{Slerp}(z_1, z_2; \mu) = \frac{\sin{\theta(1-u)}}{\sin{\theta}}z_1 + \frac{\sin{\theta\mu}}{\sin{\theta}}z_2
\end{equation}
where $\theta$ is the angle between $z_1$ and $z_2$ and $\mu \in [0,1]$.

\begin{figure}[htp]
    \centering
    \subfloat[Daily max temperature]{%
      \includegraphics[clip,width=\columnwidth]{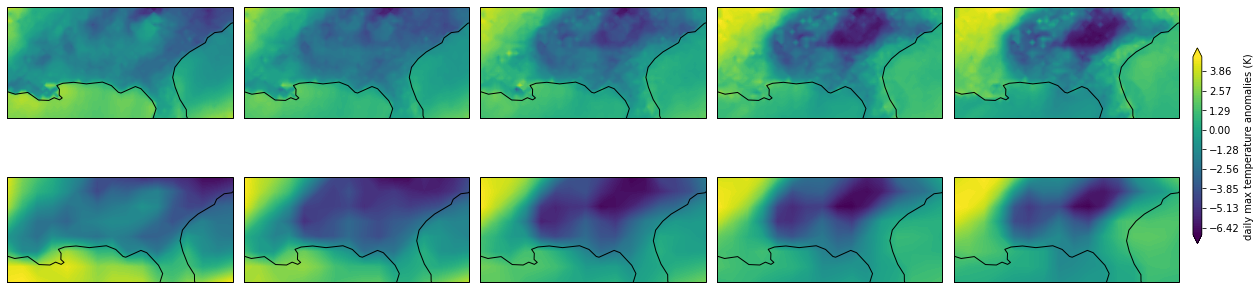}%
    }
    
    \subfloat[Precipitation]{%
      \includegraphics[clip,width=\columnwidth]{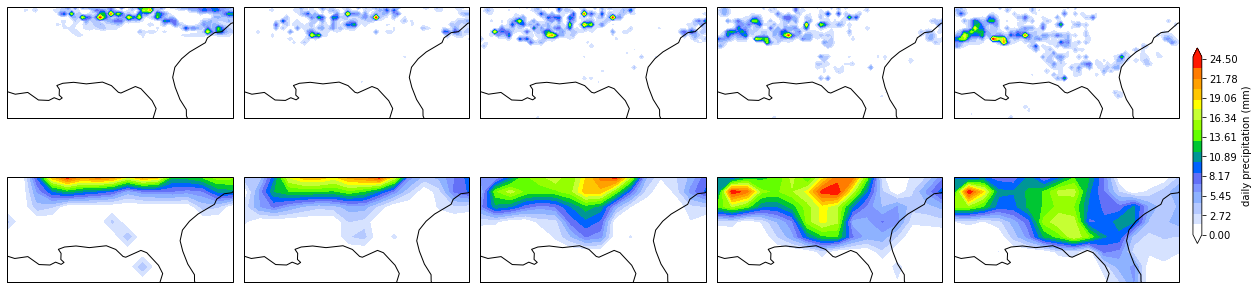}%
    }
    \caption{Temporal interpolation via the latent space, ERA-I $\rightarrow$ WRF-4; Actual WRF-4 samples (top left, top right), predicted ERA-I samples (bottom)}
    \label{fig:interp}
\end{figure}

Figure \ref{fig:interp} shows the results of applying this interpolation to the mappings learned by ClimAlign for max temperature and precipitation respectively. In both figures, the top left and top right images are real, high resolution samples from WRF-4, i.e. $y_1$ and $y_2$. The other images in the top row are the high resolution outputs produced by evaluating $f_Y$ on the interpolated values $z_k$ between $z_1$ and $z_2$. The bottom rows show the corresponding low resolution samples produced by ClimAlign for each interpolated $z_k$ as well as the two end points. For both temperature and precipitation, the figure demonstrates that interpolation in the latent space results in a smooth transition between the two real samples, with only the relevant spatial features being affected.

\subsection{Temporal misalignment}

One potential benefit of unsupervised statistical downscaling is the ability to train on \textit{temporally misaligned} datasets. For example, climate models are often used to simulate future climate conditions for a wide range of different scenarios. Assuming a sufficient level of similarity between past and future climate, we can potentially use methods such as ClimAlign to learn a mapping between variables in different time periods. The latent space $\mathcal{Z}$ can then be interpreted as shared structure which is invariant with respect to time. We consider the specific use case for statistical downscaling where our low resolution data is from a future (or simply more recent) time period than our high resolution data.

\begin{table*}[t!]
\centering
\caption{Test statistics, NOAA20CR/Livneh (2000-2013); Daily precipitation, Pacific NW}
\begin{tabular}{|c|ccc|ccccc|}
    \hline
    Method & RMSE & Bias & Corr & CDD & CWD & Rx1d & Rx5d & SDII \\
    \hline
    BCSD & 5.70 & \textbf{0.32} & 0.54 & \textbf{0.61} & \textbf{0.58} & 0.51 & 0.62 & 0.48 \\
    \hline
    ClimAlign (ours) & \textbf{5.11} & 1.04 & 0.54 & 0.47 & 0.48 & \textbf{0.53} & 0.62 & \textbf{0.49} \\
    \hline
\end{tabular}
\label{tab:misalign_prediction_error}
\end{table*}

We test this hypothesis using the NOAA 20th century reanalysis \cite{slivinski2019towards} (NOAA20CR) and Livneh \cite{livneh2013long} observation datasets, which unlike the ERA-I/WRF data, are mostly independent from each other. NOAA20CR is a reanalysis product spanning much of the 19th, 20th, and early 21st centuries. It provides global, $1^{\circ}\times 1^{\circ}$ gridded estimates of several meteorological variables based on the NCEP GFS model and surface pressure observations over land. In this work, we use a subset of the NOAA20CR dataset consisting of temperature and precipitation estimates over CONUS from 1981-2015. The Livneh dataset is a collection of daily, gridded (approx. $\frac{1}{16}^{\circ}$) meteorological and derived hydrometeorological observations for a variety of variables over CONUS and southern Canada, dating back to the early 20th century. We use a subset of the Livneh data regridded to $\frac{1}{8}^{\circ}$ as the downscaling target.

ClimAlign is trained using NOAA20CR precipitation data from 2000 through 2013 as the low resolution climate variable and Livneh observation data from 1987 through 1999 as the high resolution target\footnote{We use the same hyperparameter settings from the ERA-I/WRF experiments, with the exception of the number of levels, which is increased to 4 to account for the higher resolution. Ideally, hyperparameters should be re-tuned for each dataset.}. The model is tested on the same 2000-2013 time period but instead using the matching Livneh observations as the ground truth for evaluation. BCSD is trained on aligned NOAA/Livneh data from the 1987-1999 time period and used as a benchmark for comparison.

Table \ref{tab:misalign_prediction_error} shows the pointwise error metrics and monthly Climdex correlations computed over the test data for both methods. ClimAlign achieves lower RMSE but incurs significantly higher bias, likely due to the difference in climatology between the two time periods (2000-2013 has less precipitation, on average). Table \ref{tab:misalign_climatology} shows the bias between the predicted and true climatology for both methods. Note that P50, P98, and P02 refer to the 50th, 98th, and 2nd percentiles over wet days, respectively. While BCSD achieves better error on the mean, ClimAlign provides a better estimate of all three quantiles. This seems to indicate that it does a better job of capturing the distribution of precipitation magnitudes over the test period but not occurrence. ClimAlign also shows inferior performance on the consecutive dry/wet days indices in table \ref{tab:misalign_prediction_error} which supports this hypothesis.

\begin{table}[t!]
\centering
\caption{Climatological bias, NOAA20CR/Livneh, Daily precipitation, Pacific NW}
\begin{tabular}{|c|cccc|}
    \hline
    Method & Mean & P50 & P98 & P02 \\
    \hline
    BCSD & \textbf{0.23} & -2.37 & -4.19 & -1.05 \\
    \hline
    ClimAlign (ours) & 0.53 & \textbf{0.41} & \textbf{-3.46} & \textbf{0.01} \\
    \hline
\end{tabular}
\label{tab:misalign_climatology}
\end{table}

\begin{figure}[t!]
    \centering
    \includegraphics[width=\columnwidth]{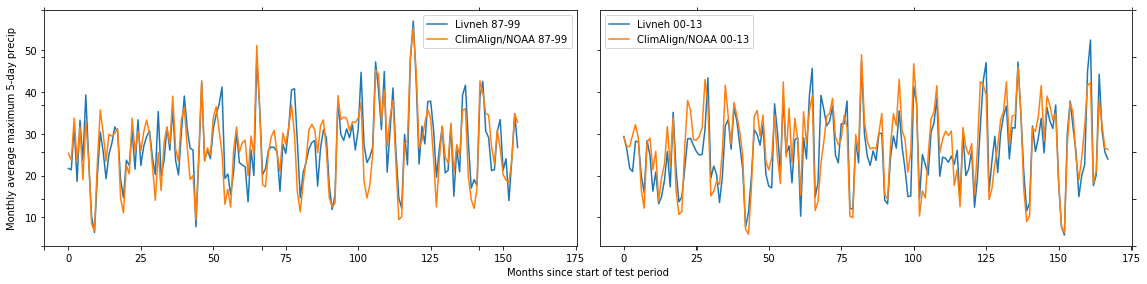}
    \caption{Monthly average maximum 5-day precipitation (Rx5d) over both time periods}
    \label{fig:climalign_noaa_livneh_series}
\end{figure}

Figure \ref{fig:climalign_noaa_livneh_series} shows monthly maximum 5-day (Rx5d) precipitation averaged over space for both the 1987-1999 (left) and 2000-2013 (right) time periods. The orange lines are the ClimAlign predictions given NOAA20CR data from each respective time period. The same fitted model parameters were used for both cases. The predictions for both time periods roughly follow the corresponding observations as expected. This indicates that ClimAlign is able to learn a reasonably consistent mapping for both time periods despite being trained on temporally misaligned data.

\section{Conclusions and future work}

In this work, we presented ClimAlign, a novel method for statistical downscaling based on recent work on a class of deep learning algorithms called \textit{normalizing flows} \cite{rezende2015variational,kingma2018glow,grover2019alignflow}. ClimAlign treats statistical downscaling as a \textit{domain alignment} problem, where the goal is to learn a mapping between two random variables based on their shared structure. To the best of our knowledge, ClimAlign is the first proposed method for \textit{unsupervised} statistical downscaling, i.e. downscaling without the use of paired low/high resolution training data. Our method also permits both unconditional and conditional sampling, allowing us to generate an arbitrary number of plausible high resolution samples for a given low resolution input. We evaluated ClimAlign on daily max temperature and precipitation data from two different datasets and demonstrated that it is capable of achieving competitive predictive performance with supervised methods despite its lack of supervision in training. We also showed that ClimAlign is able to learn a mapping between low and high resolution samples even when the training data for each resolution are from different time periods.

While ClimAlign shows promise in the application of deep, unsupervised learning to statistical downscaling, there is still much work to be done on further developing this approach. Open problems include the need to deal with the non-stationarity of climate, the incorporation of auxiliary variables, and the application of ClimAlign to other problem domains in climate science. For example, this approach may be useful in areas such as paleoclimate analysis, model parameterization, and analog mapping.

Despite having much room for improvement, ClimAlign sets a new benchmark in the application of modern machine learning methods to statistical downscaling. It provides a powerful, scalable mechanism for learning complex spatial probability distributions through state-of-the-art methods in deep generative modeling. We believe that this approach opens up a promising direction of study for both researchers and practitioners in climate informatics.

\section*{Acknowledgments}
This work was supported by research credits from Google Cloud. All experiments were carried out using the Pangeo \cite{hamman2018pangeo} software environment deployed on Google Kubernetes Engine. Datasets and technical support were provided by Jupiter Intelligence, Inc.

\bibliographystyle{ieeetr}
\bibliography{ci_references}

\end{document}